%% file: neurips_2026.tex
\documentclass{article}

\usepackage[preprint]{neurips_2026}
\author{
Jeonghwan Kim$^{1}$, Yushi Lan$^{2}$, Yongwei Chen$^{1}$, Hieu Trung Nguyen$^{3}$, \\ 
\textbf{Chuanyu Pan$^{3}$, Xingang Pan$^{1}$} \\
$^1$Nanyang Technological University, $^2$University of Oxford, $^3$Meshy AI \\
\url{https://jhkim0759.github.io/projects/SceneConductor}
}

\usepackage{multirow}
\usepackage{array}
\usepackage[utf8]{inputenc} 
\usepackage[T1]{fontenc}    
\usepackage{hyperref}       
\usepackage{url}            
\usepackage{booktabs}       
\usepackage{amsfonts}       
\usepackage{nicefrac}       
\usepackage{microtype}      
\usepackage{xcolor}         
\usepackage{booktabs}
\usepackage{pifont}
\newcommand{\cmark}{\ding{51}} 
\newcommand{\xmark}{\ding{55}}
\newcommand{\model}{SceneConductor}

\usepackage{wrapfig}
\usepackage{amsmath}
\usepackage{mathtools}
\usepackage{pifont}
\usepackage{multicol}
\usepackage{multirow}
\usepackage{makecell}
\usepackage{colortbl}
\usepackage{wrapfig}
\usepackage{float}

\title{\model: 3D Scene Generation from a Single Image with Multi-Agent Orchestration}

\begin{document}
\maketitle
\input{tex/0_abstract}

\input{tex/1_introduction}
\input{tex/2_related_work}
\input{tex/3_proposed_method}

\input{tex/4_experiments}

\input{tex/5_conclusion}
{
    \small
    \bibliographystyle{unsrt}
    \bibliography{main}
}


\end{document}

%% file: tex/0_abstract.tex
\begin{abstract}
Generating complete 3D scenes from a single image requires inferring globally consistent geometry, object relationships, and environmental context from inherently ambiguous visual evidence. Despite recent progress in joint layout-and-mesh generation, existing methods often rely on holistic or weakly decomposed pipelines that entangle many factors at once and demand extensive scene-level supervision, limiting their generalization to complex real-world environments.
We propose a multi-agent orchestration framework that performs single-image 3D scene generation through three structured stages: scene initialization, environment construction, and multi-agent refinement. The initialization stage extracts image-derived object masks, builds object-level 3D representations, and predicts an initial spatial layout to form a coarse 3D scene. The environment-construction stage then leverages this initialization together with point-map geometry to build an environmental scaffold of supporting surfaces, room boundaries, materials, and illumination. Finally, in the refinement stage, a planner agent identifies structural and visual inconsistencies, applies simple corrections directly, and dispatches specialist agents for complex localized revisions that are reintegrated into the global scene.
To provide reliable structural initialization while reducing reliance on scene-level annotations, we further introduce a geometry-aware layout predictor supervised by sparse geometric priors derived from point maps. Unlike fully supervised layout generators, the predictor can be trained from segmentation-level data and generalizes robustly to diverse real-world scenes. Extensive experiments on benchmark datasets show that our method consistently outperforms prior approaches in geometric accuracy, spatial consistency, and perceptual realism.
\end{abstract}

%% file: tex/1_introduction.tex
\section{Introduction}

3D scene generation serves as a foundation for immersive and interactive applications such as AR/VR, robotics, and digital content creation. Recent advances in generative models have substantially improved object-level 3D asset generation~\cite{mescheder2019occupancy, xie2019pix2vox, nichol2022point, vahdat2022lion, gao2022get3d, jun2023shap}, enabling high-quality reconstruction and synthesis of individual objects from images or text. Generating a complete 3D scene from a single image, however, is substantially challenging: it requires jointly inferring object geometry, spatial layout, inter-object relationships, and room-level environmental context from limited visual evidence.

Recent diffusion-based approaches address scene-level generation by jointly producing multiple objects and their layouts within a unified generation process~\cite{huang2025midi, lin2025partcrafter, meng2025scenegen, wang2026scenetransporter}. While effective at modeling global structure, maintaining separate object-level representations causes memory and computation to grow with the number of objects, making them difficult to scale to cluttered real-world scenes. Object-centric alternatives instead generate individual objects independently and compose them into a scene~\cite{sam3dteam2025sam3d, yin20263dfixer, siddiqui2026shaper}; however, their quality hinges on layout or pose modules typically trained on synthetic or domain-restricted indoor datasets~\cite{fu20203dfuture, fu20213dfront} with limited categories and canonical room configurations. As a result, both families struggle to generalize to in-the-wild images with diverse viewpoints, object compositions, and environmental structures.

On the other hand, agent-based methods offer an alternative by using large language models to plan and assemble scenes~\cite{noh2026editasact, yang2025sceneweaver, sun20253dgeneralists}, but most are driven by textual descriptions and rely on the learned priors of knowledge rather than image-grounded constraints. As a result, while they can synthesize plausible scenes, they struggle to follow image-specific guidance such as object composition, spatial arrangement, and room-level appearance. Image-conditioned multi-agent systems~\cite{yin2026viga} improve visual grounding, but typically use coarse generator/evaluator roles that operate on the entire scene; such holistic revision requires repeated multi-turn interactions over the full scene context, increasing token usage and latency while limiting the precision of localized corrections.

In this paper, we propose \model, a multi-agent orchestration framework that processing single-image 3D scene generation into three structured stages: scene initialization, environment construction, and multi-agent refinement. Unlike holistic agent-based pipelines that repeatedly reason over the entire scene, \model ~assigns each stage to specialized agents with task-specific prompts, allowing them to operate only on the relevant context required for their roles. This specialization reduces unnecessary context exposure and enables each agent to act as an expert for its assigned subtask. Moreover, rather than asking agents to generate a complete scene from scratch, we provide a strong image-grounded initialization using a geometry-aware layout predictor trained on geometric priors extracted from real-world 2D datasets.

The initialization stage extracts object-level masks from the input image, builds object-level 3D representations, and predicts an initial spatial layout to form a coarse 3D scene. The environment-construction stage uses this initialization together with point-map geometry to add room-level elements such as floors, walls, supporting structures, materials, and illumination. Finally, in the refinement stage, a planner agent inspects the scene, applies simple corrections directly, and dispatches specialist agents for complex localized revisions; each specialist focuses only on its assigned objects or region, refines them in context, and reintegrates the result into the global scene.

To further support this initialization, our geometry-aware layout predictor is designed to provide reliable layout initialization without requiring costly scene-level annotations. Instead of learning a generative scene distribution from restricted 3D datasets, it estimates object layouts from sparse geometric priors derived from segmentation masks and point maps. This allows the predictor to be trained using segmentation-level data while grounding its predictions in image-derived geometry. As a result, it yields a lightweight initialization that scales with the number of objects and generalizes to diverse real-world environments.
Our contributions are summarized as follows:
\begin{itemize}
    \item We propose a three-stage 3D scene generation framework based on multi-agent orchestration, consisting of scene initialization, environment construction, and multi-agent refinement. By assigning each agent a specialized role with structured guidelines, the framework improves task coordination and enables more reliable scene generation.
    \item We introduce a geometry-aware layout predictor that learns spatial information from real-world images using sparse geometric priors derived from segmentation masks and point maps, without requiring scene-level layout annotations, thereby improving layout initialization for subsequent scene generation.
    \item We demonstrate consistent improvements over existing approaches on benchmark datasets and real-world images in terms of geometric accuracy and perceptual quality.
\end{itemize}

%% file: tex/2_related_work.tex
\section{Related Work}
\label{sec:related_works}

\paragraph{3D Scene Generation}
3D scene generation builds on progress in 3D visual perception and object-level asset synthesis. In 3D perception, traditional structure-from-motion methods~\cite{schonberger2016structure, wang2024vggsfm} rely on iterative optimization, whereas recent feed-forward approaches~\cite{wang2024dust3r, leroy2024mast3r, wang2025vggt, chen2025easi3r, team2025aether, wang2025moge} enable efficient reconstruction from sparse or monocular observations. Their outputs are geometric observations rather than complete, editable mesh-based scenes, which limits their direct use in interactive applications.
%
In parallel, object-level 3D generation has advanced rapidly through feed-forward reconstruction~\cite{hong2023lrm,xu2024instantmesh} and diffusion-based mesh generation~\cite{liu2024one,chen2023fantasia3d,lan2024ln3diff,lan2024ga,xiang2024structured,chen2025primx,trellis,li2025triposg,chen2024sar3d,yang2024hunyuan3d2.0}, enabled by large-scale 3D asset datasets~\cite{objaverse,objaverseXL}. These methods focus on individual objects and do not address scene-level reasoning such as spatial layout and inter-object relationships.
Recent compositional 3D generative methods extend object-level synthesis by jointly modeling multiple objects, parts, or instances within a unified generation process~\cite{lin2025partcrafter, huang2025midi, meng2025scenegen, wang2026scenetransporter, chen2025ultra3defficienthighfidelity3d, ding2026fullpart, yan2025xpart, tang2024partpacker}, introducing multi-instance or part-aware representations to capture compositional structure and cross-object relationships. However, maintaining component-level representations and modeling their interactions causes memory and computation to grow with the number of objects or parts; global or cross-component attention further amplifies this cost, making it difficult to scale to cluttered real-world scenes.
A complementary line of work adopts object-centric scene construction. SAM3D~\cite{sam3dteam2025sam3d} jointly reconstructs object geometry, texture, and layout from a single image. Other image-grounded pipelines first extract image masks or object proposals, generate individual 3D assets, and then assemble them via layout prediction, scene-graph reasoning, depth or point-map alignment, or optimization-based refinement~\cite{sautter20253dregen, shi2025scenemaker, yin20263dfixer, siddiqui2026shaper, chu2023buol, yu2025metascenes, yao2025cast, dong2025hiscene, tang2025towards, zhai2023commonscenes, gu2025artiscene, ling2025scenethesis}. Although these decoupled pipelines alleviate the scalability burden of joint generation, their final quality depends heavily on the layout or pose module, which is typically trained on synthetic or domain-restricted data and therefore generalizes poorly to in-the-wild images.
In contrast, our method decouples object reconstruction from geometry-aware layout prediction, enabling scalable image-grounded scene construction without requiring scene-level annotations.

\paragraph{Agent-based Methods}
LLM/VLM-based agent frameworks have recently emerged as a flexible paradigm for 3D scene generation, leveraging language models for layout planning, asset retrieval, tool invocation, and scene editing~\cite{feng2023layoutgpt, yang2024holodeck, liu2025agentic, hunyuanworld2025tencent, sun20253dgeneralists, huang2024blenderalchemy}. While these methods can synthesize plausible scenes, they often fail to enforce reference-image constraints such as object composition, spatial arrangement, object scale, and room-level appearance.
Recent agentic systems further incorporate iterative feedback and tool-based interaction. VIGA~\cite{yin2026viga} introduces an image-conditioned dual-agent framework in which a generator and an evaluator iteratively refine scenes through execution and verification in Blender. SceneWeaver~\cite{yang2025sceneweaver} adopts a tool-augmented framework where agents repeatedly invoke specialized tools for scene construction and correction, and Edit-As-Act~\cite{noh2026editasact} reduces parameter ambiguity in tool-based editing through semantically grounded actions. While these methods improve controllability, they typically rely on coarse multi-functional agent roles that operate on the entire scene, so each revision requires multi-turn interactions over the full scene context and limits the precision of localized corrections.
In contrast, our \model~orchestrates agents with fine-grained specialized roles, enabling more effective, accurate, and structured 3D scene generation despite the inherent ambiguity of single-view input.

%% file: tex/3_proposed_method.tex
\input{figures/pipeline}

\section{Proposed Method}
Unlike prior approaches~\cite{yin2026viga, yang2025sceneweaver} that assign many responsibilities to a few agents operating holistically over the entire scene, we present a hybrid orchestration pipeline for reconstructing and refining a 3D indoor scene from a single image.
Our method integrates specialized agents with both existing models and newly proposed components, using multi-agent orchestration to mitigate the ambiguity inherent in single-view reconstruction.
The pipeline is organized into three sequential stages---scene initialization, environment construction, and multi-agent refinement---each of which incrementally improves geometric consistency and visual realism while preserving coherence with the input image (Fig.~\ref{fig:pipeline}). To provide a strong initialization and facilitate subsequent refinement, we further introduce a geometry-aware layout predictor that produces a generalizable initial scene layout (Sec.~\ref{subsec:predictor}).

\subsection{Scene Initialization}
\label{subsec:initialize}
We adopt a hybrid initialization strategy in which agents intervene selectively to improve the reliability of the initial scene representation. We first extract segmentation masks with Grounded-SAM~\cite{ren2024groundedsam}, which typically contain redundant, fragmented, or inconsistent instance boundaries. An agent refines these masks using pairwise overlap, category consistency, and spatial compatibility: highly overlapping masks with the same label are suppressed or consolidated, fragmented masks of the same object are merged, and masks covering multiple distinct objects are split. The result is a clean mask set in which each mask corresponds to a single physical object (Fig.~\ref{fig:pipeline}(a)).

From the refined masks, we reconstruct 3D object meshes with SAM3D~\cite{sam3dteam2025sam3d}, which is robust to occlusion and partial observation; for objects identified as identical by the agent, a single mesh is reconstructed and replicated to avoid redundant generation. The global scene structure is then estimated with our geometry-aware layout predictor, detailed in Sec.~\ref{subsec:predictor}.


\subsection{Environment Construction}
Given the initial scene and the predicted point map~\cite{wang2025moge}, this stage builds an environmental context---room boundaries, surface appearance, and illumination---that acts as a global spatial anchor for subsequent refinement.
At its core is a scene-aware floor-plan estimation module that establishes a reliable geometric reference frame. We reason in the bird's-eye-view (BEV) space, using both the input image and the point map to reduce perspective ambiguity.
We extract the valid 3D points from the point map as shown Fig.~\ref{fig:pipeline}(b). Projecting these points onto the horizontal plane and computing their convex hull yields a coarse \emph{stage polygon} (red, middle of Fig.~\ref{fig:pipeline}(b)) that approximates the observable spatial footprint and defines an initial environment boundary.

Combining the stage polygon with the initial object placements, the agent infers a simplified stage structure---a floor/support plane and surrounding boundary walls (right of Fig.~\ref{fig:pipeline}(b)). This is not intended to recover the full room geometry; it is a lightweight geometric scaffold that keeps objects within the valid scene extent, enforces support relationships with the floor, and prevents physically implausible placements such as floating objects or wall penetrations.
%
To further enhance realism, the agent analyzes the input image to set wall materials matching the scene's dominant color and appearance, and configures light sources for plausible illumination.

\subsection{Multi-agent Refinement}
We refine the scene through a multi-agent framework that splits the work into simple deterministic operations and targeted sub-scene optimization (Fig.~\ref{fig:pipeline}(c)).
We first extract structured scene information from both the input image and the current scene. One agent reads the input image and segmentation masks to recover object-level attributes and relational groups, while another parses the current 3D scene for spatial configurations and occlusion relationships. Conditioned on the input image and the rendered scene, a planner agent then decides the overall refinement strategy and routes each issue to a simple or a complex operation.

\paragraph{Simple operations.}
For simple inconsistencies, the planner emits a deterministic operation list that covers three cases: enforcing physical support constraints, aligning scale and orientation among similar objects, and correcting physically implausible poses or irregular scales. Because the agent specifies object targets and operation types rather than explicit numerical values, no iterative refinement is needed.

\paragraph{Complex operations.}
For tightly coupled object interactions, we adopt an isolation-based refinement strategy: the planner groups interdependent objects into a sub-scene and designates a representative anchor object as reference. A separate scene is constructed for this group, where a single specialist agent refines the configuration through multi-turn interaction with planner-provided guidance. At each iteration, the specialist inspects the newly rendered image and its segmentation, then updates the sub-scene configuration to resolve identified spatial inconsistencies. Although this refinement of spatial relationships remains iterative, its localized scope greatly reduces overhead by limiting the number of objects under consideration. The refined sub-scenes are then reintegrated into the global scene, yielding improved geometric consistency, physical plausibility, and alignment with the input image.

\subsection{Geometry-aware Layout Prediction}
\label{subsec:predictor}
\input{figures/layout_predictor}

Given the meshes and masks from Scene Initialization (Sec.~\ref{subsec:initialize}), our geometry-aware layout predictor estimates object layouts directly from the given inputs (Fig.~\ref{fig:predictor}). Unlike generative formulations, the model regresses layout parameters, which makes it trainable with non-annotated datasets.

The input image, mask, and point map are encoded as conditioning features, and each object mesh is voxelized into a low-resolution grid (e.g., $16^3$) and embedded into a compact latent through~\cite{trellis}. Learnable tokens are then updated through transformer layers with global attention module that captures inter-object interactions. The output object tokens predict per-object rotation, translation, and scale. The floor-rotation is obtained by averaging predictions across all floor-rotation tokens.

Formally, for each object $i$, the pose token predicts layout parameters $(R_i, T_i, S_i)$ in a floor-aligned canonical scene frame whose ground plane is the $XZ$-plane and whose vertical axis is $Y$. The object rotation $R_i$ is a yaw rotation around the vertical axis in this canonical frame. The global floor rotation $F$ is shared across all objects and maps the canonical floor frame to the camera coordinate frame. Given a mesh vertex $\mathbf{x}_k$ of object $i$, the final camera-frame vertex is
\begin{equation}
\hat{\mathbf{x}}_k
=
F
\left(
R_i \left( S_i \odot \mathbf{x}_k \right)
+
T_i
\right),
\label{eq:camera_transform}
\end{equation}
where $\odot$ denotes element-wise scaling.
This decomposition explicitly separates global floor alignment from per-object pose prediction: $F$ captures the floor frame's orientation in the camera, while $(R_i, T_i, S_i)$ describes each object within the canonical floor-aligned scene. Restricting object rotations to the ground plane and sharing $F$ across objects reduces the effective degrees of freedom relative to independent 6DoF prediction, enforces a consistent notion of height, and decouples object orientation from global alignment, leading to more stable and coherent layouts.
To generalize beyond limited 3D annotations, we further introduce a pointmap-supervised geometry loss that supports training without ground-truth poses, opening the door to large-scale 2D segmentation datasets.

\paragraph{Pointmap-supervised Geometry Loss.}
Let $\mathcal{M}_i = \{ \mathbf{x}_k \}$ denote points sampled from the mesh of object $i$, and $\mathcal{P}_i$ the subset of the input point map associated with object $i$. We transform $\mathcal{M}_i$ into the camera frame via Eq.~\ref{eq:camera_transform}, yielding $\hat{\mathcal{M}}_i$.
Because the point map is sparse and observes only one side of each object, we supervise with a one-sided Chamfer distance from $\mathcal{P}_i$ to $\hat{\mathcal{M}}_i$, combined with a bounding-box alignment term:
\begin{equation}
\mathcal{L}_{\text{geo}}^i =
\underbrace{
\frac{1}{|\mathcal{P}_i|} 
\sum_{\mathbf{p} \in \mathcal{P}_i}
\min_{\hat{\mathbf{x}} \in \hat{\mathcal{M}}_i}
\| \mathbf{p} - \hat{\mathbf{x}} \|_2^2
}_{\text{one-sided chamfer distance}}
+ 
\lambda 
\underbrace{
\left(
\| \hat{\mathbf{b}}_i^{xy} - \mathbf{b}_i^{xy} \|_1 
+ | \hat{\mathbf{b}}_i^{z_{\min}} - \hat{\mathbf{b}}_i^{z_{\min}} |
\right)
\vphantom{\sum_{\mathbf{p} \in \mathcal{P}_i}}
}_{\text{bounding box loss}} ,
\end{equation}
where $\mathbf{b}_i = (\mathbf{b}_i^{xy}, \mathbf{b}_i^{z_{\min}})$ is the ground-truth 3D bounding box of object $i$, with $\mathbf{b}_i^{xy}$ the ground-plane extent and $\mathbf{b}_i^{z_{\min}}$ the near-depth bound; the predicted box $\hat{\mathbf{b}}_i$ is computed from $\hat{\mathcal{M}}_i$. We supervise only near-depth direction ($z_{\min}$)) because the point map captures only visible surfaces and set $\lambda = 0.1$ to downweight the bounding-box term, since bounding boxes derived from partial observations under occlusion may not provide reliable guidance.

%% file: figures/pipeline.tex
\begin{figure}[t]
    \centering
    \includegraphics[width=1.0\textwidth]{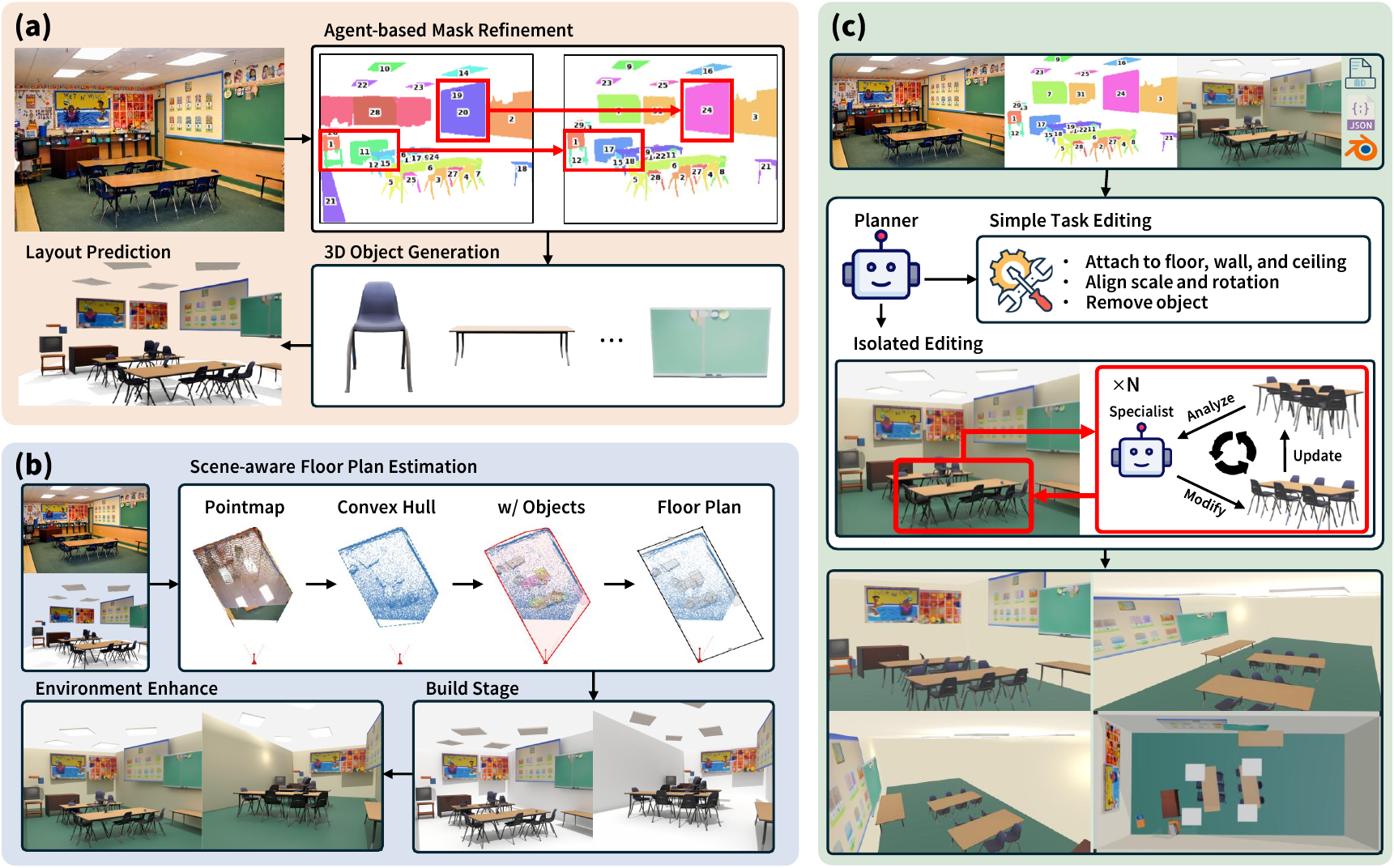}
    \vspace{-3mm}
    \caption{Overview of the \model~framework.
    We sequentially perform (a) scene initialization, (b) environment construction, and (c) multi-agent refinement to construct and refine a realistic and consistent 3D scene from a single image.
    }
    \vspace{-3mm}
    \label{fig:pipeline}
\end{figure}

%% file: figures/layout_predictor.tex
\begin{figure}[t]
    \centering
    \includegraphics[width=1.0\textwidth]{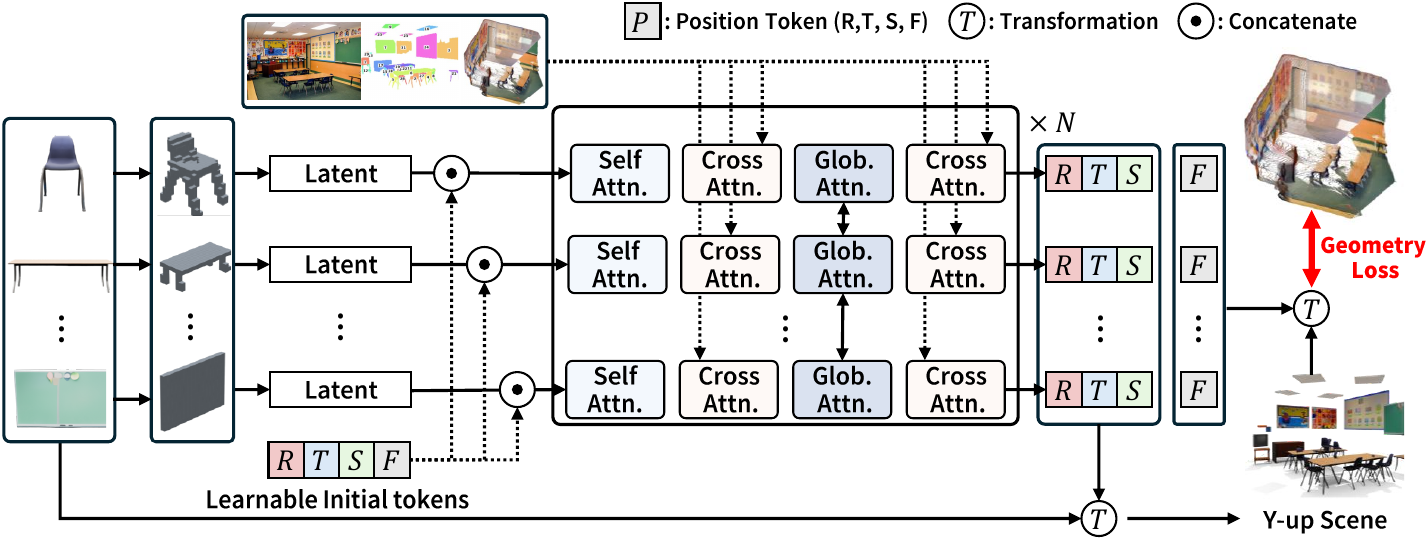}
    \vspace{-5mm}
    \caption{Overall structure of the geometry-aware layout predictor. The model takes object instances as input, together with image, mask, and point-map features, and predicts object poses parameterized by $R$, $T$, $S$, and $F$, corresponding to rotation, translation, scale, and floor rotation, respectively. These parameters define a scene aligned with the input point map.}
    \vspace{-3mm}
    \label{fig:predictor}
\end{figure}

%% file: tex/4_experiments.tex
\section{Experiments}
\input{tables/comparision}
\input{tables/mit67}
\subsection{Implementation detail}
We train the geometry-aware layout predictor using 3D-FUTURE~\cite{fu20203dfuture}, ScanNet~\cite{dai2017scannet}, and COCO~\cite{lin2014coco}. For ScanNet, pseudo labels are generated through aligning with objects, while COCO samples are filtered and trained using geometry supervision only. Since ScanNet have z-up floor structures and camera extrinsics, we can extract the ground truth floor rotation by computing the camera rotation.
Our framework follows a skill-based agent architecture, where callable functions are orchestrated by the agent, inspired by recent code-generation agent systems~\cite{claude,codex}. Furthermore, we employ Qwen3.5-27B~\cite{qwen35} for image understanding, SAM3D~\cite{sam3dteam2025sam3d} for object generation, and MoGE~\cite{wang2025moge} for point map estimation. All refine components operate within Blender and Python code, enabling direct generation of editable Blender-ready scenes. Further details on dataset construction, training protocols, and implementation settings are provided in the Supplementary Material.

\subsection{Quantitative Evaluation}
We first evaluate the performance of our geometry-aware layout prediction module against recent scene generation and refinement methods, including SceneGen~\cite{meng2025scenegen}, 3D-Fixer~\cite{yin20263dfixer}, and SAM3D~\cite{sam3dteam2025sam3d}. For quantitative evaluation, we use the 3D-FUTURE~\cite{fu20203dfuture} test set and the ScanNet~\cite{dai2017scannet} validation set with pseudo labels generated by MetaScenes~\cite{yu2025metascenes}. For ScanNet, we focus on scenes with relatively complex object compositions and refine the masks using Grounded-SAM~\cite{ren2024groundedsam} to obtain higher-quality object masks than those provided by the dataset. We further evaluate on the MIT-Indoor-67 dataset~\cite{quattoni2009mit67} to assess visual quality across diverse scene types, where object masks are extracted using Grounded-SAM due to the lack of object annotations, and report the visual metrics in Tables~\ref{tab:comparision} and~\ref{tab:mit67}.
For datasets with layout annotations, we compare the predicted scenes with the corresponding ground-truth layouts using Chamfer Distance (CD), F-score, and bounding box IoU (IoU-B). In addition to geometric metrics, we assess visual consistency using a VLM-based evaluator (i.e., Qwen3.5-2B~\cite{qwen35}), which is prompted to score the results in terms of realism, functionality, layout consistency, and image alignment, along with the CLIP~\cite{CLIP} score.
Note that the VLM metric reported in Table~\ref{tab:comparision} corresponds to the average of these VLM scores with single rendered image. Detailed evaluation settings are described in the Supplementary Material.

Table~\ref{tab:comparision} shows that our method consistently outperforms all baselines on the 3D-FUTURE dataset across both geometry and visual metrics, achieving the lowest Chamfer Distance and the highest F-Score and IoU-B. These results indicate more accurate and structurally consistent 3D reconstructions, while the improvements in VLM and CLIP-S suggest better alignment with visual semantics. Table~\ref{tab:mit67} provides complementary evidence from perceptual evaluations on the MIT-Indoor-67 test set, where our method achieves the best performance across all criteria, including realism, functionality, layout, and image alignment. Notably, the gains in layout and image alignment highlight the effectiveness of our geometry-aware design in producing coherent object arrangements. Overall, these results demonstrate that our approach improves both geometric fidelity and perceptual quality, validating the effectiveness of geometry-aware layout prediction.

\input{tables/multi-agent}
To evaluate the complete orchestration framework, we compare our full pipeline results with VIGA~\cite{yin2026viga}, a representative multi-agent approach for scene generation. 
We randomly sample 20 scenes from the MIT-Indoor-67 dataset and compare the results in Table~\ref{tab:multi-agent}. Note that this setting yields lower scores than the previous evaluations because the object-focused prompt (``Focus only on object-level correspondence to the reference image.'') is not used; instead, the VLM evaluates the entire scene context and therefore penalizes scene-level inconsistencies more heavily.
As shown in Table~\ref{tab:multi-agent}, our full pipeline outperforms VIGA across the evaluated metrics, demonstrating the effectiveness of our specialized multi-agent orchestration. Moreover, compared with our initialization-only results, the full pipeline achieves higher overall scores, indicating that the environment construction and planner-guided refinement stages further improve scene fidelity and structural consistency.

\input{figures/3dfuture}

\input{figures/mit67}

\subsection{Qualitative Evaluation}
To demonstrate the effectiveness of our geometry-aware layout prediction, we compare our method with SceneGen and SAM3D on the 3D-FUTURE and ScanNet datasets, as illustrated in Fig.~\ref{fig:3dfuture}. As can be seen, our method predicts more accurate object rotations than prior approaches, resulting in better alignment across objects and producing layouts that are more consistent with the input image. Specifically, for samples from ScanNet scenes where the camera viewpoint is significantly tilted, competing methods often generate unstable layouts in which individual objects lie on inconsistent ground planes. In contrast, our method preserves a coherent and parallel ground-plane structure across objects. This improvement arises from our training strategy, which explicitly enforces alignment between the xz-plane and the physical floor. Consequently, the proposed representation not only establishes a normalized and geometrically consistent 3D layout, but also provides a standardized structural format that supports more reliable downstream processing within the orchestration framework.

Furthermore, we compare the outputs of each stage in our orchestration process—initialization, environment construction, and multi-agent refinement—with VIGA~\cite{yin2026viga} in Fig.~\ref{fig:mit67}. The results show that our approach produces more realistic and well-composed scenes than VIGA while achieving superior layout fidelity. Although VIGA employs a multi-agent framework based on generator–evaluator interactions, it still often struggles to faithfully reflect the input scene, even when allowed up to 30 rounds of dialogue. This limitation arises from its holistic scene-level revision strategy: agents repeatedly reason over the entire scene context, resulting in substantial communication overhead and making precise localized corrections challenging.
In contrast, our framework begins with strong initialization, providing a reliable foundation for subsequent refinement and reducing the need for complex scene-level planning from scratch. The refinement stage further decomposes corrections into simple direct edits and complex isolated editing, enabling more precise and realistic scene improvements, such as fixing floating objects and aligning objects with the wall, as shown in the fourth column of Fig.~\ref{fig:mit67}.

\section{Discussion}
\input{tables/ablation}
\input{figures/ablation}
\paragraph{Ablation Study}
We evaluate the performance of the geometry-aware layout prediction on the ScanNet dataset, as reported in Table~\ref{tab:ablation}. We compare models with progressively incorporated components, including the geometry loss, floor rotation, and training with the segmentation dataset. We observe that while incorporating floor rotation alone does not consistently improve geometric metrics, it stabilizes pose prediction by reducing rotational ambiguity. This effect becomes more evident when combined with segmentation data, leading to the best overall performance.
We further compare qualitative results across different variants in Fig.~\ref{fig:ablation}. With geometry loss, object scales and spatial extents are estimated more accurately, leading to improved geometric fidelity. Incorporating floor rotation results in more consistent and aligned object orientations with respect to the ground plane (xz-plane). Finally, when all components are incorporated, the model achieves the best overall performance, producing more accurate scale and translation estimates as well as more coherent object placement.

\paragraph{Limitations}
Our framework relies on multiple foundation models (e.g., Qwen3.5, SAM3D, and MoGE) executed sequentially, which introduces non-trivial inference latency and errors from early stages may also propagate to later refinement steps. In addition, our current experiments mainly focus on indoor scenes with texture-level appearance modeling. Extending the framework to outdoor or unbounded environments and richer material representations remains an important direction for future work.
%

%% file: tables/comparision.tex
\begin{table}[t]
    \centering  
    \caption{Quantitative comparison of scene reconstruction metrics on 3D-FUTURE and ScanNet.}
    \label{tab:comparision}
    \begin{tabular}{ll|ccc|cc}
    \toprule
      &   & \multicolumn{3}{c}{Geometry Metrics} & \multicolumn{2}{c}{Visual Metrics} \\
    \cmidrule(lr){3-5} \cmidrule(lr){6-7}
    Dataset & Model
    & CD $\downarrow$ & F-Score $\uparrow$ & IoU-B $\uparrow$ 
    & VLM $\uparrow$ & CLIP-S $\uparrow$ \\
    \midrule
    \multirow{4}{*}{3D-FUTURE}
    & 3D-Fixer     & 0.0378 & 0.8133 & 0.4165 & 6.2297 & 0.7738 \\
    & SceneGen     & 0.0137 & 0.8603 & 0.4426 & 8.2208 & 0.8029 \\
    & SAM3D        & 0.0117 & 0.8928 & 0.5100 & 8.3805 & 0.8046 \\
    & Ours & \textbf{0.0089} & \textbf{0.9261} & \textbf{0.5735} & \textbf{8.5140} & \textbf{0.8071} \\
    \midrule
    \multirow{4}{*}{ScanNet}
    & 3D-Fixer     & 0.1713      & 0.4761     & 0.1909      & 2.0773 & 0.6512 \\
    & SceneGen     & 0.1915     & 0.4204      & 0.1655      & 3.3850 & 0.6724\\
    & SAM3D        & 0.1717     & 0.4544      & 0.1789      & 3.2300 &  0.6482 \\
    & Ours         & \textbf{0.1511}     & \textbf{0.5206}     & \textbf{0.2080}     & \textbf{4.7633} & \textbf{0.6734} \\
    \bottomrule
    \end{tabular}
\end{table}

%% file: tables/mit67.tex
\begin{table}[t]
\centering
\caption{Comparison of MIT-Indoor-67 test set with visual metric.}
\label{tab:mit67}
\begin{tabular}{lcccccc}
\toprule
Method 
& Realism $\uparrow$ 
& Functionality $\uparrow$ 
& Layout $\uparrow$ 
& Image-Align $\uparrow$ 
& Avg. $\uparrow$ 
& CLIP $\uparrow$ \\
\midrule
SceneGen  & 3.0280 & 2.4703 & 2.5239 & 2.0898 & 2.5280 & 0.6138 \\
3D-Fixer  & 2.6305 & 1.9437 & 1.9725 & 1.7747 & 2.0804 & 0.5903 \\
SAM3D     & 4.3283 & 4.5776 & 3.7741 & 3.3916 & 4.0179 & 0.6236 \\
Ours      & \textbf{4.5665} & \textbf{4.8839} & \textbf{4.0200} & \textbf{3.6264} & \textbf{4.2742} & \textbf{0.6302} \\
\bottomrule
\vspace{-2mm}
\end{tabular}
\end{table}


%% file: tables/multi-agent.tex
\begin{table}[t]
\centering
\caption{Comparison of methods on 20 samples from the MIT-Indoor-67 test set.}
\label{tab:multi-agent}
\resizebox{\linewidth}{!}{%
\begin{tabular}{lcccccc}
\toprule
Method 
& Realism $\uparrow$ 
& Functionality $\uparrow$ 
& Layout $\uparrow$ 
& Image-Align $\uparrow$ 
& Avg. $\uparrow$ 
& CLIP $\uparrow$ \\
\midrule
SAM3D                & 2.050 & 1.200 & 1.250 & \textbf{2.200} & 1.675 & 0.751  \\
VIGA                 & 1.800 & 0.900 & 1.100 & 1.400 & 1.300 & 0.750  \\
Initialization       & 2.250 & 1.450 & 1.250 & 2.100 & 1.762 & 0.717 \\
SceneConductor       & \textbf{2.400} & \textbf{1.600} & \textbf{1.400} & 2.150 & \textbf{1.887} & \textbf{0.779} \\
\bottomrule
\end{tabular}%
}
\vspace{-1mm}
\end{table}

%% file: figures/3dfuture.tex
\begin{figure}[t]
\vspace{-4mm}
    \centering
    \includegraphics[width=1.0\textwidth]{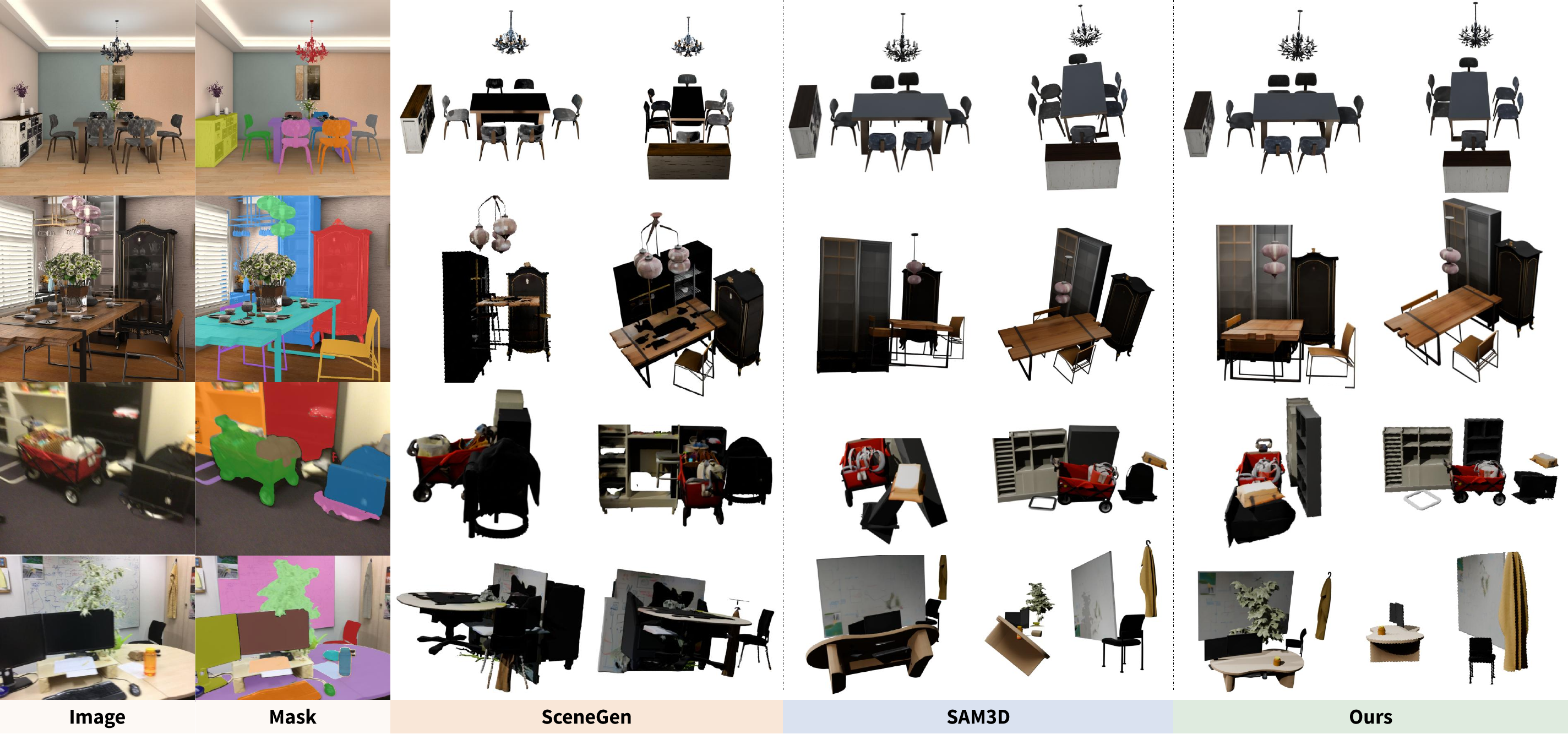}
    \vspace{-4mm}
    \caption{Qualitative comparison of our geometry-aware layout prediction against SceneGen and SAM3D. The first two rows are from 3D-FUTURE, and the last two are from ScanNet.
    }
    \label{fig:3dfuture}
\end{figure}

%% file: figures/mit67.tex
\begin{figure}[t]
\vspace{-2mm}
    \centering
    \includegraphics[width=1.0\textwidth]{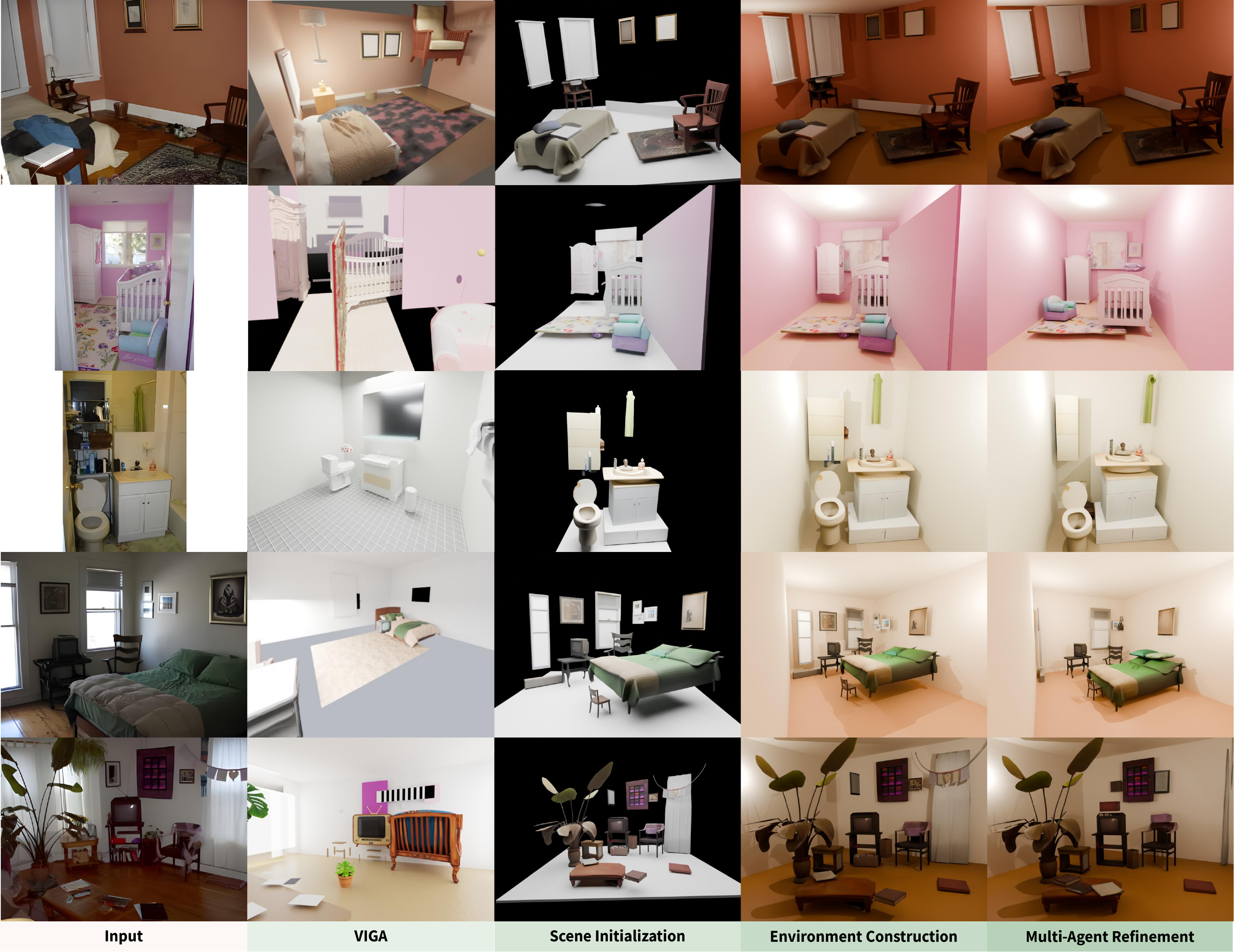}
    \vspace{-5mm}
    \caption{Qualitative comparison of each stage in our \model ~ and VIGA on MIT-Indoor-67.}
    \label{fig:mit67}
\end{figure}

%% file: tables/ablation.tex
\begin{table}[t]
\centering
\caption{Ablation study on key components in ScanNet validation set}
\label{tab:ablation}
\begin{tabular}{ccc|ccc|cc}
\toprule
\multicolumn{3}{c|}{Components} 
& \multicolumn{3}{c|}{Geometry Metrics} 
& \multicolumn{2}{c}{Visual Metrics} \\
\cmidrule(lr){1-3} \cmidrule(lr){4-6} \cmidrule(lr){7-8}
Geometry & Floor Rot. & Seg. Data 
& CD $\downarrow$ & F-Score $\uparrow$ & IoU-B $\uparrow$ 
& VLM $\uparrow$ & CLIP-S $\uparrow$ \\
\midrule
\xmark & \xmark & \xmark 
& 0.1530 & 0.4943 & 0.2036 & 4.5933 & 0.6704 \\

\cmark & \xmark & \xmark 
& 0.1618 & 0.4817 & 0.1903 & 4.4583 & 0.6708 \\

\cmark & \cmark & \xmark 
& 0.1595 & 0.4857 & 0.1950 & 4.4550 & 0.6717 \\

\cmark & \cmark & \cmark 
& \textbf{0.1511} & \textbf{0.5206} & \textbf{0.2080} & \textbf{4.7633} & \textbf{0.6734} \\
\bottomrule
\end{tabular}
\end{table}

%% file: figures/ablation.tex
\begin{figure}[t]
    \centering
    \includegraphics[width=1.0\textwidth]{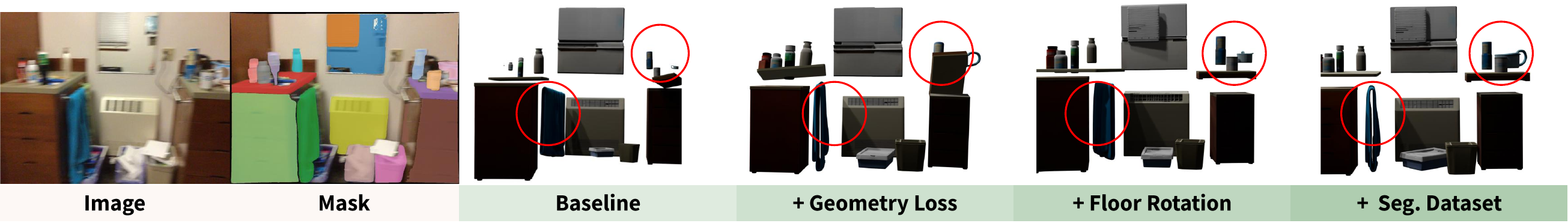}
    \vspace{-3mm}
    \caption{Qualitative ablation results. Adding geometry loss, floor rotation, and segmentation data progressively improves layout quality.}
    \label{fig:ablation}
    \vspace{-3mm}
\end{figure}    

%% file: tex/5_conclusion.tex
\section{Conclusion}
We present \model, an orchestration-based framework for generating 3D scenes from a single image. By decomposing scene generation into structured stages, our approach makes the generation process easier for agents to manage and control. We further propose a robust layout prediction method that improves both accuracy and generalization. Experiments demonstrate that \model ~outperforms prior methods. We believe orchestration offers a promising direction for scalable and controllable image-to-scene generation.